# A Mission Planning System for the AUV "SLOCUM Glider" for the Newfoundland and Labrador Shelf

M. Eichhorn* Member IEEE, C. D. Williams* Member MTS, R. Bachmayer+, Member IEEE and B. de Young†
*Institute for Ocean Technology
National Research Council Canada
Arctic Avenue, P.O. Box 12093
St. John's, Newfoundland A1B 3T5, Canada
phone +01 709-772-7986; fax + 01 709-772-2462
e-mail: mike.eichhorn@nrc-cnrc.gc.ca

+Faculty of Engineering and Applied Sciences
†Department of Physics and Physical Oceanography
Memorial University of Newfoundland
St. John's, Newfoundland A1B 3X7, Canada

*Abstract-* **This paper presents a system for mission planning for an autonomous underwater vehicle in time-varying ocean currents. The mission planning system is designed for the AUV "SLOCUM Glider" to collect oceanographic data along the Newfoundland and Labrador Shelf. The data will be used in conjunction with a numerical ocean model currently under development by the Department of Fisheries and Oceans Canada. This allows for the validation and the modification of existing ocean current and climate models as well as the design of new models with the aim of improving the accuracy of forecasts. The use of the ocean current forecast data in netCDF format in an ocean current model, the algorithms which consider glider-specific behaviour, details of the program's technical implementation in C++, and, preliminary results will be described.**

## I. Introduction

The concept presented in this article is part of a feasibility study for a future project between the National Research Council Canada's Institute for Ocean Technology, Memorial University of Newfoundland and the Department of Fisheries and Oceans (DFO) Canada [1] to collect salinity, temperature, ocean current and ice thickness data along the Newfoundland and Labrador Shelf using autonomous underwater gliders.

The goal of the mission planning system is to generate an optimized glider path plan between predefined start and end locations. The algorithm takes vehicle dynamics and the time varying ocean currents into account. By including the ocean currents in the path generation process, the glider is able to either utilize the currents to advance, or avoid regions of adverse currents that would otherwise hinder the glider when passing through that region. The route search algorithm is based on a modified Dijkstra Algorithm including the time-variant cost function in the search. This algorithm uses graph-based methods and is described in detail in [2] and [3]. The software uses the standard netCDF [4] file format to interface with ocean prediction systems in order to extract temporal and spatial ocean current information about the region of interest. For the purpose of this experiment the data are provided through the DFO's Canada-Newfoundland Operational Forecast System (C-NOOFS). The generation of fine scale ocean current information for an arbitrary position, depth and time using these netCDF data files in a multi-dimensional interpolation scheme is a crucial factor for a successful route search.

Fig. 1 shows the mission planning system with its several programming parts. Due to the limited scope of the ten day forecast window as well as differences between the offline calculated path and the real sailed mission route, it is necessary to define subordinate target positions for a possible mission plan update via satellite.

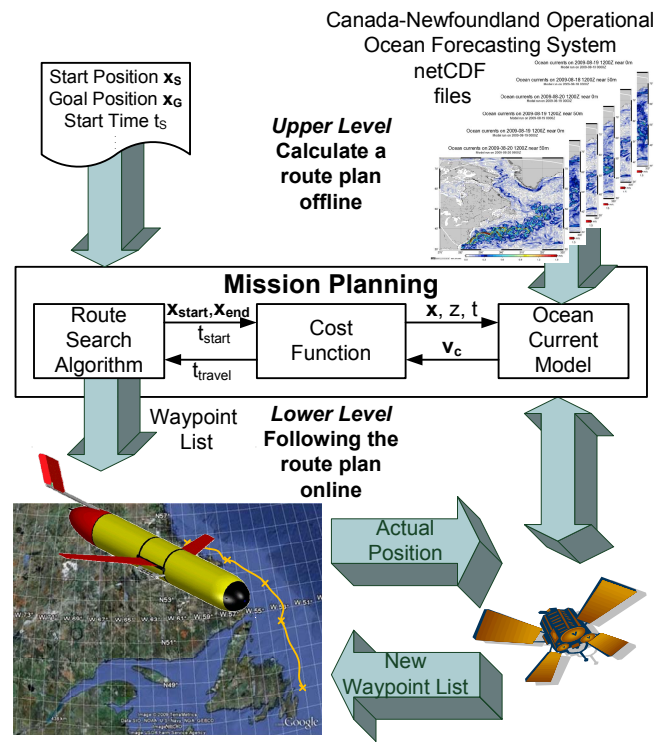

Figure 1. Mission Planning System in overview

## II. OCEAN CURRENT MODEL

The cost value for the several edges in the used geometrical graph is the time required for traversing the edge (see section III.B in [2]). This time depends on the length of the path segment and the speed $v_{path\_ef}$, at which the vehicle travels along the path in relation to the fixed world coordinate system. The speed $v_{path\_ef}$ depends on the vehicle speed through the water $v_{veh\_bf}$ (cruising speed), the amount and the direction of the ocean current vector as well as the direction of the path. (A detailed description of the cost value calculation is presented in section III in [3].) To determine the travel time, accurate ocean current information along the path element is required.

This ocean current information will be provided through the DFO's Canada-Newfoundland Operational Forecast System (C-NOOFS) in the form of netCDF files, generated in a numerical model. At present it is not possible to directly couple the search algorithm and the numerical model because of time and computational constraints. Important steps for using netCDF files in an ocean current model in the path planning system will be described below.

### A. Interface between the current model and the cost function

The cost function to calculate the travel time functions like a numerical simulation. This simulates the vehicle driving along the path from a defined start position $\mathbf{x}_{start}$ at start time $t_{start}$ to the end position $\mathbf{x}_{end}$ under consideration of the local ocean current. The result of the simulation is either the travel time or the knowledge that the path is impassable due to adverse ocean current. During such a simulation the cost function requires a large amount of local ocean current information, which is generated in the ocean current model. Thereby the Cartesian position $\mathbf{x}$, the depth $z$ and the time $t$ is passed from the cost function to the ocean current model. The model returns a two dimensional ocean current vector $\mathbf{v}_c = [u\ v]$. The software technical implementation will be described in section IV.

### B. Preparation of the netCDF-Files

The Network Common Data Format (NetCDF) is a binary data format for array-oriented scientific data [4] which is commonly used for climatology, meteorology and oceanography applications. The C-NOOFS provides the ocean current data at various depths (0 to 5700 m) for the entire Northwest Atlantic with a resolution of ca. six km in the region of interest, every six hours for a 10-day forecast in geographical coordinates. Such a data file contains 12.1 GB of data. To use these data in the ocean current model, the following modifications are necessary:

- decrease the spatial density of the data points
- use a Cartesian coordinate system as reference for the ocean current information
- reduce the data volume to a practical form, that means
- use only the depth layers of interest, through which the glider can pass (this is determined by the depth rating of glider itself)
- use only the region of interest, where the mission is planned and
- remove unnecessary oceanographic data from the files.

These six objectives can be addressed by using the FIMEX library [5] (also see section IV).

### C. Multi-dimensional interpolation scheme

Since the ocean current data coming from the forecasting system as data files, will be provided only at discrete times and positions with a coarser time and length scale than is required to generate an efficient path, a multi-dimensional interpolation scheme will be utilized to generate the desired data. Fig. 2 shows the scheme for the ocean current component $v$ in overview. The first interpolation step uses a two-dimensional interpolation function from the FIMEX library to extract the ocean current information for the several depth layers. A Nearest-Neighbour, a Bilinear and a Bicubic interpolation method are available. The interpolations via the depth and time dimensions occur separately using one-dimensional interpolation functions. Nearest-Neighbour, Linear, Cubic and Akima interpolation are possible. The first two methods require two fields (for time $t$) or layers (for depth $z$), the other methods require minimal three, optimal five fields or layers in order to generate the ocean current component $v$ at the defined position $\mathbf{x_i}$, at the depth $z_i$ and at the time $t_i$. The implementation of the Akima interpolation [6] can make allowance for an abrupt change of ocean current conditions in case of tides or of different depth streams. The curves which are created with this method do not have unnatural wiggles in the case of an abrupt change in value, which occurs when the Cubic interpolation method is used.

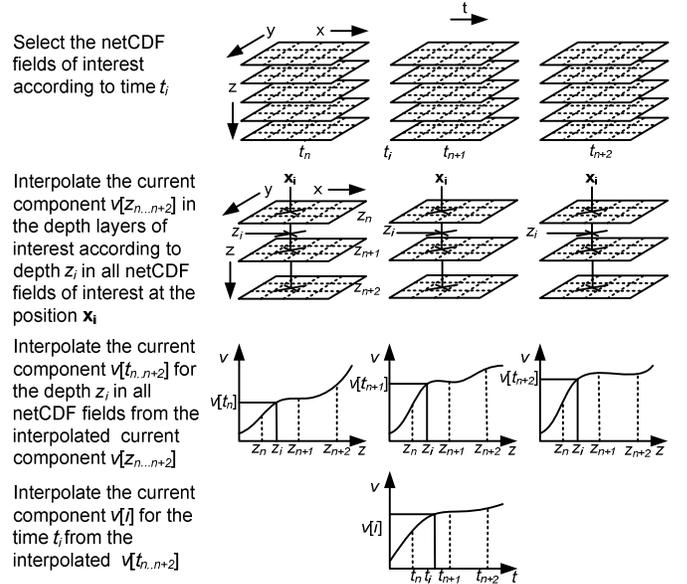

Figure 2. Steps to interpolate an ocean current component $v[i]$ at position $\mathbf{x}[i]$, depth $z_i$ and time $t_i$ from the netCDF fields

## III. DETAILS OF THE ALGORITHMS

This section describes the algorithms required to include glider-specific behaviors in the mission planning system, specifically the prescribed dive profiles and restrictions to accurately follow the waypoints.

### A. Glider dive profile cost function

The glider dive profile is specified by its locomotion principle. By changing its buoyancy, the glider is able to descend (dive) and ascend (climb). The result is a saw-tooth vertical profile as shown schematically in Fig. 3. The exact simulation [7] of such a dive profile is computationally time-intensive and so is impractical because of the number of edges in the geometrical graph, which range from one hundred thousand to one million. Conversely, the knowledge of the glider's behaviour in every passable depth is necessary for the planning and makes it possible that the mission planning can avoid regions with an adverse surface or seabed current.

To include the depth-varying ocean current information in the cost function (presented in section III in [3]) the path element is divided into several path segments. The number of the segments $n_{segments}$ is defined by the step size $h$. This number shall consider the changeable ocean current conditions along the path at every passable depth. In each segment the glider dives from the "climb-to" depth $z_{climb-to}$ until the "dive-up" depth $z_{dive-up}$; see Fig. 4. The calculated travel time $t_{travel}$ for the segment will correspond approximately to the travel time which the glider needs to travel along every segment of the saw-tooth profile. Table 1 includes the details of the algorithm to calculate the travel time. Fig. 3 shows the simplified dive profile in comparison to the real saw-tooth profiles.

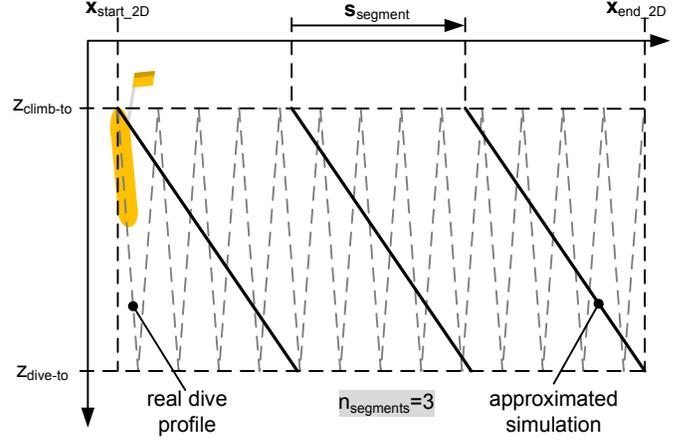

Figure 3. Simplified dive profile along a path element

### B. Optimal Dive Profile

Passing through regions with an adverse surface or seabed current by using a constant dive profile with possible large depth amplitude in order to collect oceanographic data at each depth, is impossible. The search algorithm will create a path to drive around these areas, taking the long way around to arrive at the goal point. Data in the region of interest will therefore not be collected. The simulation of a selection of dive profiles with different "climb-to" depths $z_{climb-to}$ and "dive-to" depths $z_{dive-up}$ distributed over the maximum permitted depth profile in every cost function can solve the problem.

The created dive profiles are specified through a minimum diving depth $z_{min}$, a maximum diving depth $z_{max}$, a maximum "climb-to" depth $z_{climb-to\_max}$, a minimum dive amplitude $z_{minRange}$, and the number of "climb-to" $n_{climb-to\_levels}$ levels and "dive-to" levels $n_{dive-to\_levels}$. Fig. 4 shows a possible dive profile selection. Every cost function calculates $n$ travel time values for the various dive profiles according the algorithm in Table 1. The profile with the least travel time provides the cost value which is used. It is also possible to use the travel time generated from the dive profile with the maximum depth amplitude, if the travel time is less than a defined maximum. Therefore the requirement to collect as much oceanographic data as possible during a mission can be included.

TABLE I
PSEUDO-CODE OF THE ALGORITHM TO CALCULATE THE TRAVEL TIME FOR THE GLIDER ALONG A DEFINED DIVE PROFILE

TRAVELTIME-GLIDER($x_{start\_2D}$, $x_{end\_2D}$, $z_{climb-to}$, $z_{dive-to}$, $t_{start}$)
*defined parameters: $h$*
**if** ($t_{start} = \infty$)
   **return** ($t_{travel} = \infty$)
$n_{segments} = \text{ceil}(1/h)$
$s_{segment\_2D} = (x_{end\_2D} - x_{start\_2D})/n_{segments}$
$t_{start\_segment} = t_{start}$
$x_{start\_segment\_x} = x_{start\_2D\_x}$
$x_{start\_segment\_y} = x_{start\_2D\_y}$
$x_{start\_segment\_z} = z_{climb-to}$
$x_{end\_segment\_z} = z_{dive-to}$
**for** ($i = 1$) **to** ($i = n_{segments}$)
   $x_{end\_segment\_x} = x_{start\_segment\_x} + s_{segment\_2D\_x}$
   $x_{end\_segment\_y} = x_{start\_segment\_y} + s_{segment\_2D\_y}$
   $t_{travel} = \text{TRAVELTIME}(x_{start\_segment}, x_{end\_segment}, t_{start\_segment})$
   **if** ($t_{travel} = \infty$)
      **return** $t_{travel}$
   $t_{start\_segment} = t_{start\_segment} + t_{travel}$
   $x_{start\_segment\_x} = x_{end\_segment\_x}$
   $x_{start\_segment\_y} = x_{end\_segment\_y}$
**return** ($t_{travel} = t_{start\_segment} - t_{start}$)

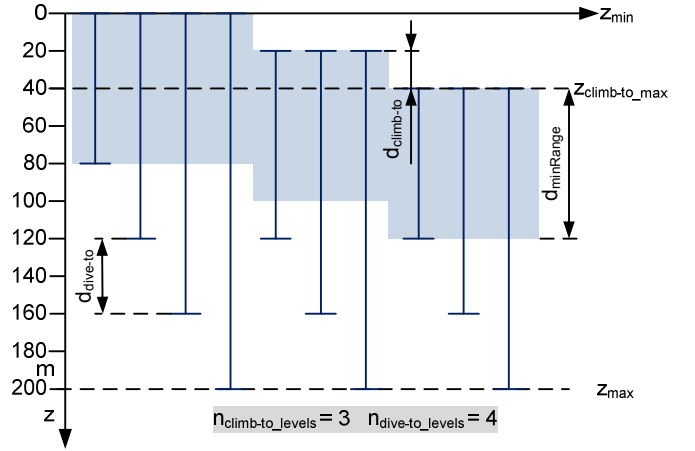

Figure 4. Created dive profiles

## C. Path smoothing

The required geometrical graph for the search algorithm is not complete as not all vertices are connected by an edge within the graph (see section III.B in [2]). A complete directed graph on *n* vertices has *n*(*n*-1) edges, which would evolve into a very large graph and a long computing time for the path search. So, the search algorithm can consider in its search only the edges which are included in the non-complete graph. The paths found are characterized by many several path segments with change of directions. The run of such a path is stair- or wiggle-shaped, which a glider cannot follow. Therefore a method to smooth the candidate path will be presented as follows.

TABLE II
ALGORITHM FOR PATH SMOOTHING IN TIME-VARYING ENVIRONMENT

| |
|---|
| PATH-SMOOTHING (*WP*, $t_0$) |
| $TT[1] = t_0$ |
| **for** (*i* = 2) **to** (*i* = length(*WP*)) |
|    $TT[i] = TT[i-1]$ + TRAVELTIME(*WP*[*i*-1], *WP*[*i*], *TT*[*i*-1]) |
| **while** (length(*WP*) ≠ length(*WP*$_{smooth}$)) AND (length(*WP*) > 2)    **I** |
|   $i_{start} = 1$ |
|   $u_{smooth} = 1$ |
|   $TT_{smooth} = \emptyset$ |
|   $WP_{smooth} = \emptyset$ |
|   $TT_{smooth}[u_{smooth}++] = TT[1]$ |
|   $WP_{smooth}[u_{smooth}++] = WP[1]$ |
|   $t_{travel\_1} = TT[2] - TT[1]$ |
|   **for** ($i_{path}$ = 3) **to** ($i_{path}$ = length(*WP*))    **II** |
|     *merge* = true |
|     $t_{travel\_2}$ = TRAVELTIME(*WP*[$i_{path}$-1], *WP*[$i_{path}$], *TT*[$i_{path}$-1]) |
|     $t_{travel\_sum}$ = TRAVELTIME(*WP*[$i_{start}$], *WP*[$i_{path}$], *TT*[$i_{start}$])    **III** |
|     **if** ($t_{travel\_sum} = \infty$) |
|       *merge* = false |
|     **else** |
|       **if** ($t_{travel\_1} + t_{travel\_2}$) < $t_{travel\_sum}$    **IV** |
|         *merge* = false |
|       **else** |
|         $t_{end} = TT[i_{start}] + t_{travel\_sum}$    **V** |
|         **for** ($i_{end} = i_{path} + 1$) **to** ($i_{end}$ = length(*WP*)) |
|           $t_{end} = t_{end}$ + TRAVELTIME(*WP*[$i_{end}$-1], *WP*[$i_{end}$], $t_{end}$) |
|           **if** ($t_{end} > TT[end]$) |
|             *merge* = false |
|           **else** |
|             $TT[end] = t_{end}$ |
|     **if** (*merge* = true) |
|       $t_{travel\_1} = t_{travel\_sum}$ |
|       $TT[i_{path}] = TT[i_{start}] + t_{travel\_sum}$ |
|     **else** |
|       $TT[i_{path} - 1] = TT[i_{start}] + t_{travel\_1}$ |
|       $TT[i_{path}] = TT[i_{start}] + t_{travel\_1} + t_{travel\_2}$ |
|       $i_{start} = i_{path} - 1$ |
|       $t_{travel\_1} = t_{travel\_2}$ |
|       $TT_{smooth}[u_{smooth}++] = TT[i_{start}]$    **VI** |
|       $WP_{smooth}[u_{smooth}++] = WP[i_{start}]$ |
|   **if** (*merge* = false) |
|     $TT[end] = TT[end-1]$ + TRAVELTIME(*WP*[end-1],*WP*[end],*TT*[end-1]) |
|   $TT_{smooth}[u_{smooth}++] = TT[end]$ |
|   $WP_{smooth}[u_{smooth}++] = WP[end]$ |
|   $TT = TT_{smooth}$ |
|   $WP = WP_{smooth}$ |
| **return** *WP* |

Table II includes the details of the algorithm to smooth the path under consideration in the time-varying environment. The candidate path is described by a list of waypoints *WP*. The algorithm verifies the start point *WP*[$i_{start}$] of the list with the subsequent waypoints *WP*[$i_{path}$] of a direct connection (**III**), with the goal of a quicker arrival at this point by using the several path elements (**IV**). Verification of the arrival time *TT*[*end*] of the goal point *WP*[*end*] from the tested waypoint *WP*[$i_{path}$] using the existing subsequent waypoints (**V**) also occurs. A second verification through the time-varying environment is necessary and ensures that the merge of path elements indeed leads to a quicker arrival at a local waypoint, but leads to a later arrival time at the goal point. This is possible even through the ocean current situation is changing dynamically. The merging begins by the third waypoint (**II**) and will be executed until one of the two verifications is satisfied or the goal point is reached. In the case of a positive verification (*merge = false*), the present waypoint *WP*[$i_{path}$] will be stored in the new waypoint list and a new merge will begin at the precedent waypoint (**VI**). The result is a waypoint list *WP*$_{smooth}$ with fewer waypoints in the verified waypoint list *WP*. This above described procedure will be repeated until the number of waypoints is constant between two sequent loops (**I**). Fig. 5 shows an example for a better understanding.

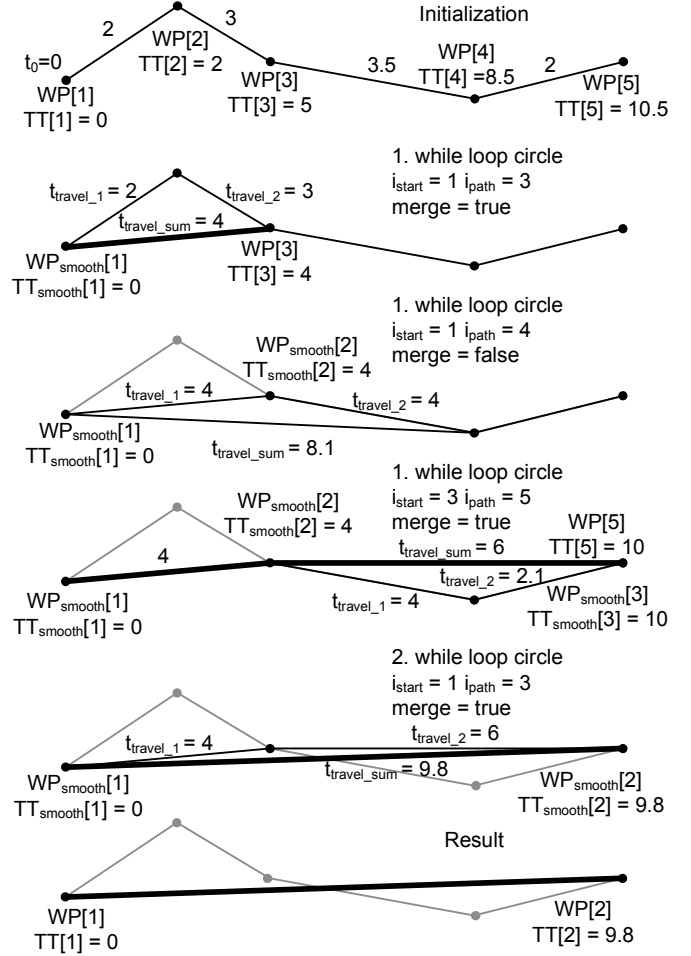

Figure 5. Several steps to smooth a path

## IV. SOFTWARE TECHNICAL DETAILS

This section provides an overview of some software technical details, and, the software products and libraries used. Fig. 6 provides an overview of the various components of the mission planning system with the software used.

The mission planning system is programmed in C++ using the Microsoft Visual Studio 2005 [15]. The data which include the ocean current information of a ten day forecast is available in netCDF file format (see section II.B). In order to use these data for mission planning, a converter program using the FIMEX library is written. FIMEX is a File Interpolation, Manipulation, and Extraction library for gridded geospatial data, written in C/C++. It supports different data formats (currently netCDF, NcML, grib1/2 and felt), to convert from one format into another, to change the projection and interpolation of scalar and vector grids as well as to subset the gridded data and to extract only portions of the files [5].

This converter also allows the use of one of the above-listed data formats in the mission planning process. So there exists a universal interface for a great number of oceanographic data formats for the mission planning system. The output of the converter provides the required input formats for the mission planning. These are netCDF data files which include the ocean current information in Cartesian coordinates. The projection in Cartesian coordinates is necessary, because all calculations in the mission planning system are based on a Cartesian coordinate system.

An XML file, which includes all necessary mission parameters such as start position, goal position, start time, region of interest, restricted areas, obstacles, projection definitions (projection function, reference position, cartographic parameters) and dive profile parameters (see section III.B) are passed to the converter and to the mission planning system.

The Xerces-C++ XML Parser [9] is used to parse the files in several C++ programs. The Boost Graph Library (BGL) [10, 11] is used in the route planning algorithm. It contains a library with complete data structures for graphs and search algorithms (Dijkstra, A*) and supports many C++ compilers. The visitor concepts of the BGL, which allow the users to insert their own operations in the existing graph algorithms, make it easy to implement the TVE algorithm, used for a search in a time-varying environment (TVE) (see section III.B in [2]), in the Breadth First Search algorithm. The resulting path includes a list of waypoints in Cartesian coordinates. To use this list in a real mission for the glider, the positions will be transformed into geographical coordinates using the Proj4 library [12] and will be saved in XML files using the Xerces-C++ XML Parser. Such a file includes a list of waypoints, consisting of the position in Cartesian and geographical coordinates, the arrival time at the point as well as the dive profile parameters to reach the point.

Xalan-J (Java) [13] is used to transform the XML files into the glider-specific mission planning text files for the existing operational system of the glider (GliderDOS [14]). (Xalan-Java is an XSLT processor for transforming XML documents into HTML, text, or other XML document types.) It will also be used to generate SVG documents (Scalable Vector Graphics) [15] from the XML files, which will be viewed in Microsoft Internet Explorer or in a vector graphic program. To analyze and to provide a graphical illustration of the resulting path, MATLAB will be used [16]. The conversion of data in the XML files to MATLAB data types occurs with use of the xml_io_tools library [17] and [18].

The cost function to determine the optimal dive profile for the path segments, which is presented in section III.B, allows easy and efficient parallel processing. The multiple cost functions for the defined dive profiles can be calculated independently at the same time on the separate processor cores of a multi-core computer. The Task library, which is a component of the Boost Sandbox (The Boost Sandbox contains un-reviewed code that is intended to eventually become a part of the Boost libraries [20]), provides interesting concepts and functions to solve this challenge, where a thread pool is created at the start of the mission planning system. The cost function, called in the search algorithm, passes the various functions for the dive profiles in the pool. After the calculation of all functions in the pool, the determination of the final cost value based on the calculated cost values occurs.

The importation of the netCDF files in the ocean current model occurs within the NetCDF library [4]. All ocean current data will be stored in multidimensional array structures. For the xy-interpolation of the ocean current from the several depth layers, the FIMEX library is used (see also section II.C).

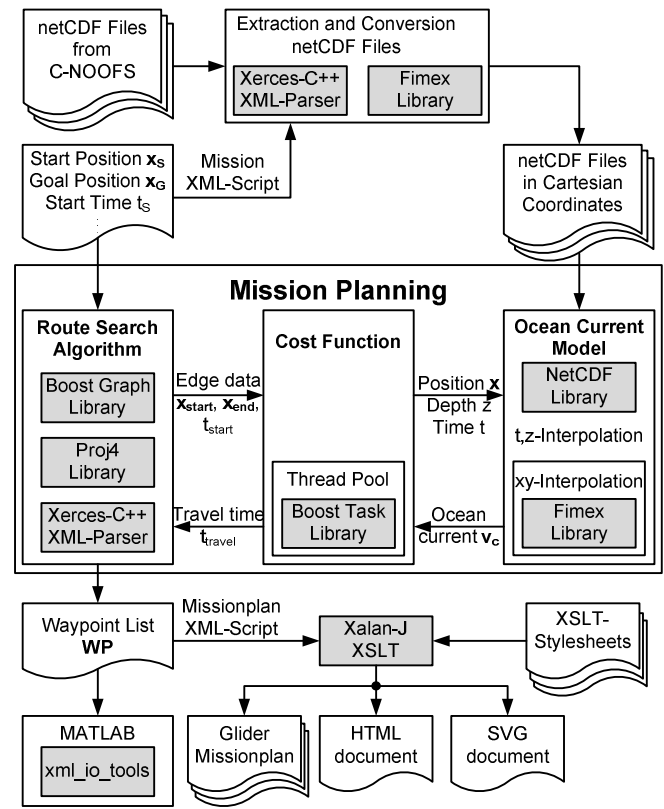

Figure 6. Software concept of the mission planning system

## V. RESULTS

This section presents the results of the mission planning system using real netCDF files for a 10-day forecast.

### A. Scaling of the netCDF files and interpolation methods

The first part of the tests shows the influence of the chosen scaling of the ocean current data in the prepared netCDF files (see section II.B) in combination with different xy-interpolation methods during the path search. Table III contains the results. Fig. 7 to Fig. 9 show the paths found using various xy-interpolation methods in the ocean current model by using ocean current data with different scaling. The different ocean current profiles in the figures illustrate the time-varying ocean flow during the mission.

TABLE III
RESULTS OF THE DIFFERENT GRID SIZES AND INTERPOLATION METHODS

| xy-Interpolation Method Grid size | Travel Time d:h:min:s | Path Length km | Comp. Time s |
|---|---|---|---|
| Nearest-Neighbour 1.0 km | 07:04:48:11 | 331.71 | 16.16 |
| Nearest-Neighbour 2.5 km | 07:06:06:02 | 338.10 | 15.50 |
| Nearest-Neighbour 5.0 km | 07:05:13:52 | 338.12 | 15.43 |
| Nearest-Neighbour 10.0km | 07:11:01:36 | 337.67 | 15.63 |
| Bilinear 1.0 km | 07:04:48:02 | 332.65 | 20.36 |
| Bilinear 2.5 km | 07:06:35:57 | 337.70 | 19.59 |
| Bilinear 5.0 km | 07:06:32:05 | 337.72 | 19.59 |
| Bilinear 10.0 km | 07:15:04:54 | 342.18 | 20.10 |
| Bicubic 1.0 km | 07:04:43:52 | 332.29 | 28.48 |
| Bicubic 2.5 km | 07:06:28:46 | 338.15 | 27.21 |
| Bicubic 5.0 km | 07:05:54:36 | 337.72 | 26.80 |
| Bicubic 10.0 km | 07:13:24:14 | 343.31 | 27.69 |

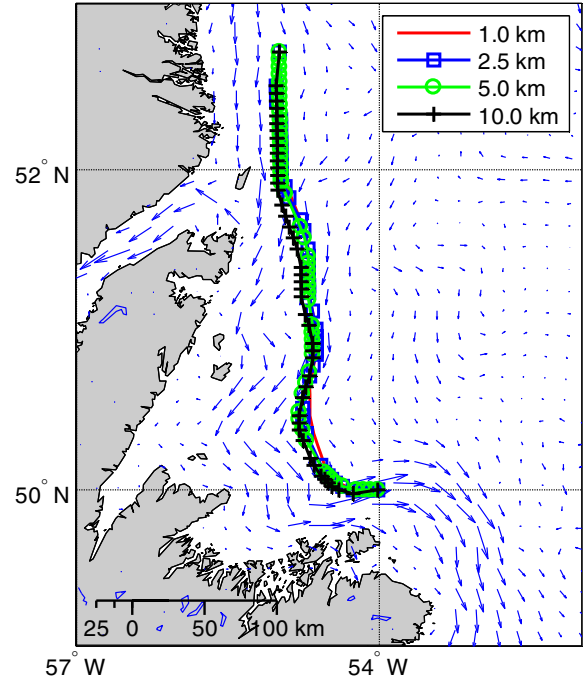

Figure 8. Time optimal path using the Bilinear xy-interpolation in the current model during the path search with different grid sizes

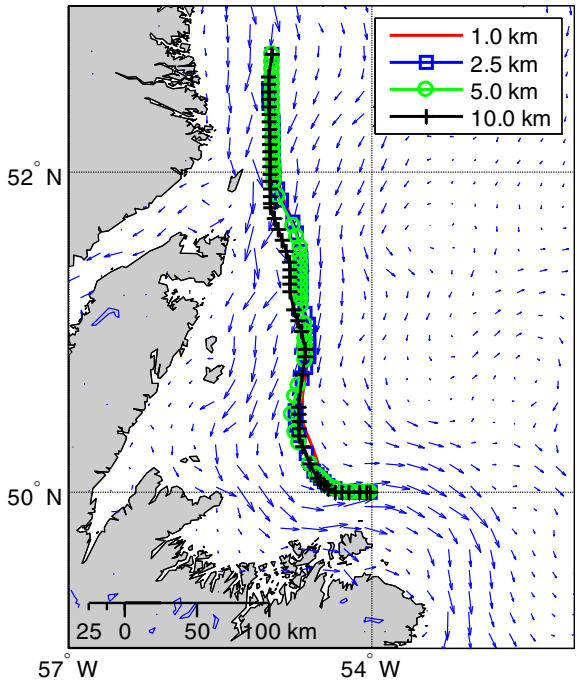

Figure 7. Time optimal path using the Nearest-Neighbour xy-interpolation in the current model during the path search with different grid sizes

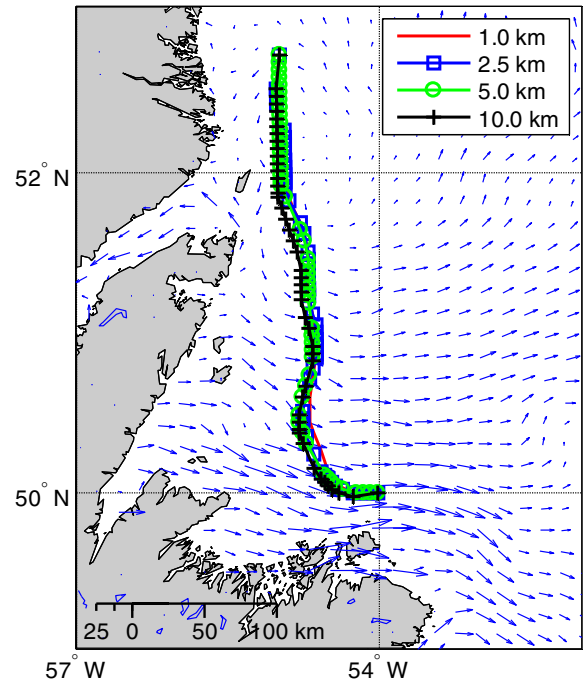

Figure 9. Time optimal path using the Cubic xy-interpolation in the current model during the path search with different grid sizes

All paths found using the different xy-interpolation methods and grid sizes have similar trajectories. The paths found using the 10 km grid size of the ocean current data vary slightly from the "main track". A decrease of the grid size does not create an appreciable improvement. The travel times as well as the path lengths of the several paths are similar. It is also clear that decreasing the grid size does not create more detailed information about the ocean current gradient than the information provided in the data files. Conversely, a refinement of the data mesh size of the prepared netCDF files for the path search allows the use of the Nearest-Neighbour or Bilinear interpolation in the ocean current model during the search without an important loss of quality in comparison to the Bicubic interpolation. This will decrease the computing time in the optimal dive profile algorithm (see section III.B) or in the algorithm used to find the optimal departure time (see section VI).

Table IV includes the results of the various interpolation methods used to determine the depth and the time values. The results are similar (see also Fig. 10) by all four methods whereas the computing times vary significantly.

TABLE IV
RESULTS OF THE DIFFERENT DEPTH-TIME INTERPOLATION METHODS

| (z,t) Interpolation Method Grid size | Travel Time d:h:min:s | Path Length km | Comp. Time s |
|---|---|---|---|
| Nearest-Neighbour 2.5 km | 07:07:07:54 | 339.14 | 6.29 |
| Linear 2.5 km | 07:06:32:23 | 338.10 | 6.48 |
| Akima 2.5 km | 07:06:28:46 | 338.15 | 27.28 |
| Cubic 2.5 km | 07:06:26:15 | 338.15 | 34.33 |

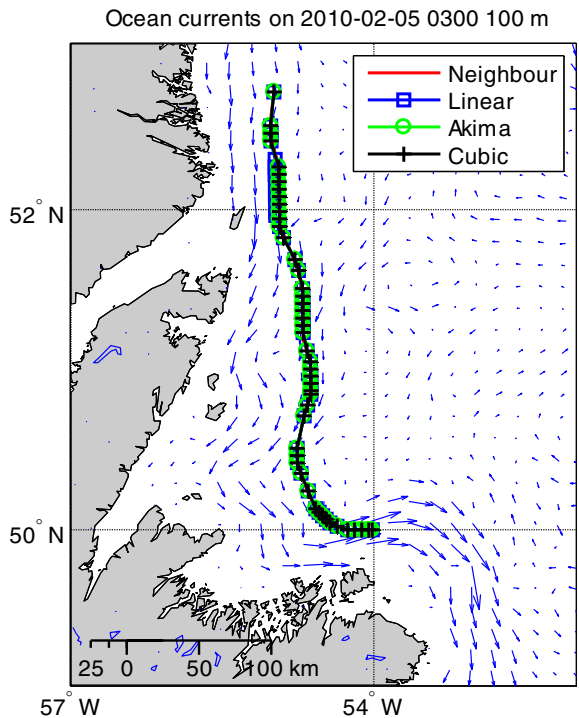

Figure 10. Time optimal path using several (z,t) interpolations in the current model during the path search

### B. Possible missions and path smoothing algorithm

This section presents the results using the path smoothing algorithm presented in section III.C by means of a selection of the possible missions along the Newfoundland and Labrador Shelf. Table V includes the results of the travel time and the length of the paths found, the smoothed path, and straight line to the goal point. Furthermore, the number of waypoints for the generated (unsmoothed) path and the smoothed path are shown. Fig. 11 shows the mission paths. The trajectories of the unsmoothed and the smoothed path of the missions are similar. The length of all missions is 210 km. The utilization of the Labrador Stream in Mission M13, M14, M23 and M24 brings a remarkable decrease of the mission endurance in comparison to the other missions. The number of waypoints in the missions can be decreased on average more than half using the smoothing algorithm, which improves the resulting path with respect to travel time. This is possible because new connections (edges) will be created which were not available in the geometrical graph during the search. An additional decrease of the waypoint list is possible, when longer travel times to the goal point are accepted (see Table II, Marker V). The direct course to the goal point leads to longer travel times or is impassable in the case of an adverse ocean current.

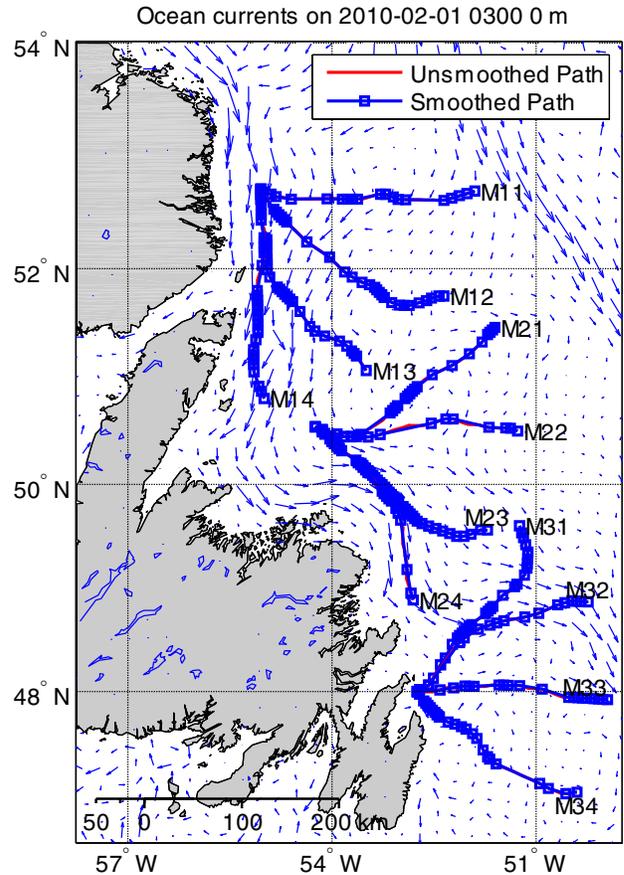

Figure 11. Time optimal path for different missions along the Newfoundland and Labrador Shelf

TABLE V
RESULTS OF THE PATH SMOOTHING ALGORITHM BY USING DIFFERENT MISSIONS

| Mission | Travel Time Unsmoothed Path d:h:min:s | Travel Time Smoothed Path d:h:min:s | Travel Time Straight Line d:h:min:s | Travel Time Straight Line without Current d:h:min:s | Path Length Unsmoothed Path km | Path Length Smoothed Path km | Path Length Straight Line km | No. of Waypoints Unsmoothed Path | No. of Waypoints Smoothed Path |
|---|---|---|---|---|---|---|---|---|---|
| Mission 11 | 08:19:06:06 | 08:17:34:14 | NaN | 08:02:26:40 | 215.87 | 215.41 | 210.00 | 64 | 19 |
| Mission 12 | 08:05:21:12 | 08:05:03:13 | 09:04:53:02 | 08:02:57:13 | 229.40 | 229.26 | 210.55 | 47 | 26 |
| Mission 13 | 05:11:44:13 | 05:11:38:42 | 06:05:45:51 | 08:02:57:13 | 228.30 | 228.01 | 210.55 | 54 | 31 |
| Mission 14 | 04:10:26:36 | 04:10:03:30 | 04:23:12:36 | 08:02:26:40 | 214.24 | 213.44 | 210.00 | 65 | 30 |
| Mission 21 | 07:23:58:18 | 07:23:47:35 | 08:19:10:56 | 08:02:57:13 | 226.91 | 226.51 | 210.55 | 46 | 23 |
| Mission 22 | 08:04:14:05 | 08:03:16:06 | 08:12:01:05 | 08:02:26:40 | 216.49 | 215.30 | 210.00 | 52 | 14 |
| Mission 23 | 05:23:38:22 | 05:23:25:01 | 07:06:46:35 | 08:02:57:13 | 224.86 | 224.86 | 210.55 | 50 | 41 |
| Mission 24 | 04:09:37:18 | 04:09:02:41 | NaN | 08:02:57:13 | 222.19 | 221.09 | 210.55 | 49 | 25 |
| Mission 31 | 08:12:03:13 | 08:11:46:59 | 10:07:59:45 | 08:02:57:13 | 229.71 | 229.41 | 210.55 | 46 | 30 |
| Mission 32 | 07:09:42:55 | 07:09:27:14 | NaN | 08:02:57:13 | 221.54 | 221.18 | 210.55 | 38 | 16 |
| Mission 33 | 08:03:48:33 | 08:02:24:53 | 08:12:11:46 | 08:02:26:40 | 213.25 | 211.74 | 210.00 | 68 | 19 |
| Mission 34 | 07:12:12:10 | 07:12:03:51 | 07:23:04:58 | 08:02:57:13 | 221.01 | 220.50 | 210.55 | 39 | 23 |

## C. Varying of the glider speed

To investigate the influence of the glider speed on the planned paths, Mission 3 in section B will be used. Furthermore this variation allows the inclusion of inaccuracy in the path planning as a result of forecast error variance, accuracy of calculation in the cost functions and a different vehicle speed in the real mission than planned. The speed is varied from 0.25 to 0.35 m/s. A comparison of the results for the missions is presented in Table VI. The shaded text fields highlight that the travel time exceeds the forecast horizon of ten days. In such a case the ocean current model calculates the ocean current with the end time of the forecast. Fig. 12 shows the determined paths using various vehicle speeds.

Fig. 13 shows the magnitude of the existing ocean current and the angle between the ocean current and the path direction during mission M31 at the passed waypoints at different depths. If the ocean current is larger than the vehicle speed (see 4.-5. day), the only way for a successful maneuvering is to follow of the current direction ($|\psi|<90°$) until the glider passes beyond this area.

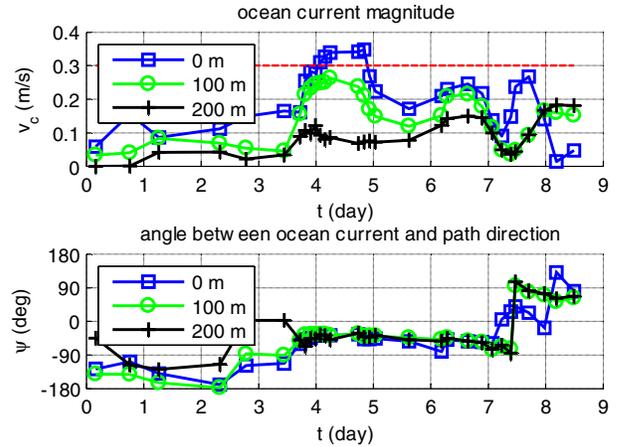

Figure 13. Ocean current magnitude at different depths for M31 $v_{veh}$ = 0.3 m/s

TABLE VI
RESULTS FOR VARIOUS VEHICLE SPEEDS

| Mission | Travel Time Path d:h:min:s | Travel Time Straight Line d:h:min:s | Path Length Path km | Path Length Straight Line km |
|---|---|---|---|---|
| M31 0.25 m/s | 11:03:03:19 | NaN | 233.89 | 210.55 |
| M31 0.30 m/s | 08:12:03:13 | 10:07:59:45 | 229.71 | 210.55 |
| M31 0.35 m/s | 07:09:13:37 | 08:05:53:55 | 227.94 | 210.55 |
| M32 0.25 m/s | 08:16:04:06 | NaN | 226.68 | 210.55 |
| M32 0.30 m/s | 07:09:27:14 | NaN | 221.18 | 210.55 |
| M32 0.35 m/s | 06:11:34:01 | 07:09:27:39 | 227.99 | 210.55 |
| M33 0.25 m/s | 09:12:19:43 | NaN | 214.00 | 210.00 |
| M33 0.30 m/s | 08:02:24:53 | 08:12:11:46 | 211.74 | 210.00 |
| M33 0.35 m/s | 06:18:36:39 | 07:01:42:44 | 212.08 | 210.00 |
| M34 0.25 m/s | 08:15:55:13 | 09:07:43:58 | 225.00 | 210.55 |
| M34 0.30 m/s | 07:12:03:51 | 07:23:04:58 | 220.50 | 210.55 |
| M34 0.35 m/s | 06:13:03:21 | 06:20:33:53 | 218.97 | 210.55 |

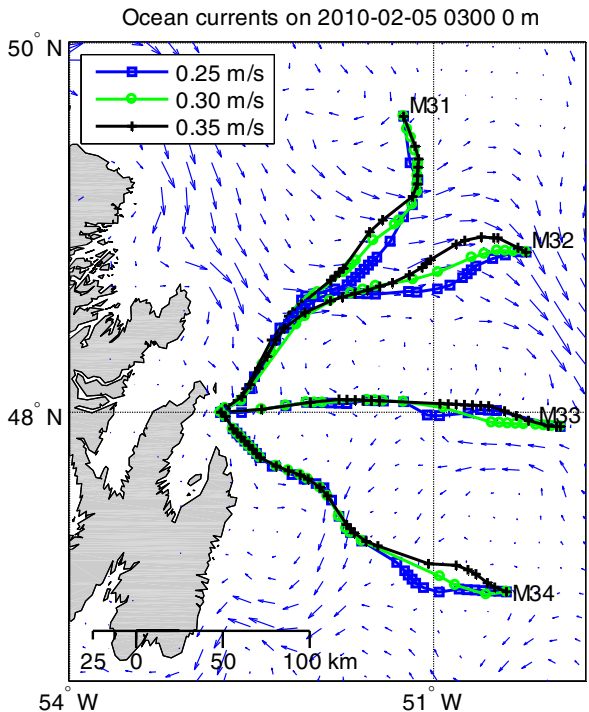

Figure 12. Time optimal path for different missions

## VI. Conclusions

In this paper a system for mission planning for an autonomous underwater vehicle in time-varying ocean currents is presented. The paper builds upon a search algorithm for a time–varying environment which is described in [2] and [3]. The first part of this paper describes an ocean current model to extract the ocean current information from netCDF data files. This model will be used in the cost function calculation during the path search. The design goal of this model was to provide as accurately as possible ocean current information with a low computing time. The fast calculation is required because the number of edges in the geometrical graph ranges from one hundred thousand to one million for a mission of duration 10 days. An algorithm to smooth the generated path is presented in the central part of this paper. Its use allows a significant decrease of the number of waypoints found in the search by concurrently advancing the travel time. A few software technical details as well as the software products and libraries used were also presented.

The influence of the scaling of the ocean current data in combination with different interpolation methods was determined by means of practical tests. All tests used real netCDF files for a 10-day forecast to analyse the feasibility of real missions along the Newfoundland and Labrador Shelf in a future collaborative project.

The presented path planning system creates a time-optimal path from a start point to a goal point by a defined departure time. A future research topic is the optimization of the departure time. The initial tests used a root-finding algorithm to detect the minimum travel time in dependence of the departure time. A symbolic wave-front expansion technique for an Unmanned Air Vehicle (UAV) or Autonomous Underwater Vehicle (AUV) in a time-varying environment (for wind or ocean current) was introduced in [21] to solve this problem. An analysis of this technique for use in the mission planning system is another research task.

For the verification of the path found in a simulation, the use of an accurate glider-model based on [7] will be investigated. This model includes the steady flying behaviour (steady dives and steady climbs) and the behaviour during three types of transitions (dive-to-climb, climb-to-dive and turning from side-to side).

The first at-sea test of the mission planning system will begin in the Summer of 2010. The goal of these tests is the validation of the offline generated mission plan with a real glider mission for the proposed test area along the Newfoundland and Labrador Shelf to the northeast of the island of Newfoundland [22].

## Acknowledgments

This work is financed by the German Research Foundation (DFG) within the scope of a two-year research fellowship. The authors would like to thank the Department of Fisheries and Oceans Canada for its support during this project, in particular Debbie Power and Dr. Fraser Davidson. Special thanks go to Dr. Heiko Klein from the Norwegian Meteorological Institute for his help with the FIMEX library.

## References

[1] Fisheries and Oceans Canada, Newfoundland and Labrador, "Canada-Newfoundland Operational Ocean Forecasting System", 2010, http://www.c-noofs.gc.ca/php/home_e.php.
[2] M. Eichhorn, "A New Concept for an Obstacle Avoidance System for the AUV "SLOCUM Glider" Operation under Ice", Oceans '09 IEEE Bremen, 2009.
[3] M. Eichhorn, "Optimal Path Planning for AUVs in Time-Varying Ocean Flows" 16th Symposium on Unmanned Untethered Submersible Technology (UUST09), 2009.
[4] Unidata, "NetCDF (network Common Data Form) Website", 2010, http://www.unidata.ucar.edu/software/netcdf/.
[5] Norwegian Meteorological Institute, "FIMEX Website", 2010, https://wiki.met.no/fimex/start.
[6] H. Akima, "A New Method of Interpolation and Smooth Curve Fitting Based on Local Procedures", Journal of the Association for Computing Machinery, vol. 17, no. 4, 1970, pp. 589-602
[7] M. He, C.D. Williams and R. Bachmayer, "Simulation of an Iterative Planning Procedure for Flying Gliders into Strong Ocean Currents" 16th Symposium on Unmanned Untethered Submersible Technology (UUST09), 2009.
[8] Microsoft MSDN, "Visual C++", 2010, http://msdn.microsoft.com/en-ca/visualc/default.aspx
[9] XERCES, "Xerces WebSite", 2010, http://xerces.apache.org/xerces-c/
[10] boost, "boost C++ Libraries", 2010, http://boost.org.
[11] J.G. Siek, L. Lee and A. Lumsdaine, *The Boost Graph Library*, Addison-Wesley, 2002.
[12] PROJ.4, "PROJ.4 Cartographic Projections library", 2010, http://trac.osgeo.org/proj/.
[13] The Apache Software Foundation, "Xalan-Java WebSite", 2010, http://xml.apache.org/xalan-j/index.html.
[14] Teledyne Webb Research, *User Manual, Slocum Glider,* 2008.
[15] W3G, "Scalable Vector Graphics (SVG)", 2010, http://www.w3.org/Graphics/SVG/.
[16] The MathWorks, "The MathWorks Homepage", 2010, http://www.mathworks.com.
[17] J. Tuszynski, "Tutorial for xml_io_tools Package", 2010, http://www.mathworks.com/matlabcentral/fx_files/12907/5/content/html/xml_tutorial_script.html.
[18] MATLAB CENTRAL, "xml_io_tools", 2010, http://www.mathworks.de/matlabcentral/fileexchange/12907-xmliotools.
[19] boost, "Boost Subversion repository Revision 60865 sandbox/task," 2010, https://svn.boost.org/svn/boost/sandbox/task/.
[20] boost, "Boost Sandbox WebSite", 2010, https://svn.boost.org/trac/boost/wiki/BoostSandbox.
[21] M. Soulignac, P. Taillibert and M. Rueher, "Time-minimal Path Planning in Dynamic Current Fields", *IEEE International Conference on Robotics and Automation,* 2009.
[22] NCOG, "Newfoundland Center for Ocean Gliders", 2010, http://www.physics.mun.ca/~glider/.